\documentclass[twocolumn]{article}
\usepackage{graphicx} 
\usepackage[margin=1in]{geometry} 
\usepackage{xcolor} 
\usepackage[normalem]{ulem} 
\usepackage[switch]{lineno} 

\makeatother\usepackage{amsmath} 
\usepackage{amssymb} 
\usepackage{bm} 
\usepackage{empheq}
\usepackage{abstract}
\usepackage{optidef}
\usepackage{float}
\usepackage{stfloats}
\usepackage{placeins}
\usepackage{hyperref}
\usepackage{authblk} 

\date{}
\begin{document}
\title{Wake Vectoring for Efficient Morphing Flight}
\author[1,2]{Ioannis Mandralis$^\ast$}
\author[1]{Severin Schumacher}
\author[1]{Morteza Gharib}
\affil[1]{\small Department of Aerospace Engineering, California Institute of Technology}
\affil[2]{\small Department of Computing and Mathematical Sciences, California Institute of Technology}

\twocolumn[
\maketitle
\vspace{-4em}
\begin{center}
{\small $^{\ast}$Corresponding author: \href{imandralis@caltech.edu}{imandralis@caltech.edu}}
\end{center}
\begin{abstract}
Morphing aerial robots have the potential to transform autonomous flight, enabling navigation through cluttered environments, perching, and seamless transitions between aerial and terrestrial locomotion. Yet mid-flight reconfiguration presents a critical aerodynamic challenge: tilting propulsors to achieve shape change reduces vertical thrust, undermining stability and control authority. Here, we introduce a passive wake vectoring mechanism that recovers lost thrust during morphing. Integrated into a novel robotic system, Aerially Transforming Morphobot (ATMO), internal deflectors intercept and redirect rotor wake downward, passively steering airflow momentum that would otherwise be wasted. This electronics-free solution achieves up to a $40\%$ recovery of vertical thrust in configurations where no useful thrust would otherwise be produced, substantially extending hover and maneuvering capabilities during transformation. Our findings highlight a new direction for morphing aerial robot design, where passive aerodynamic structures, inspired by thrust vectoring in rockets and aircraft, enable efficient, agile flight without added mechanical complexity.
\end{abstract}
]

\section*{Introduction}

\begin{figure*}[thbp]
    \centering
    \includegraphics[width=\linewidth]{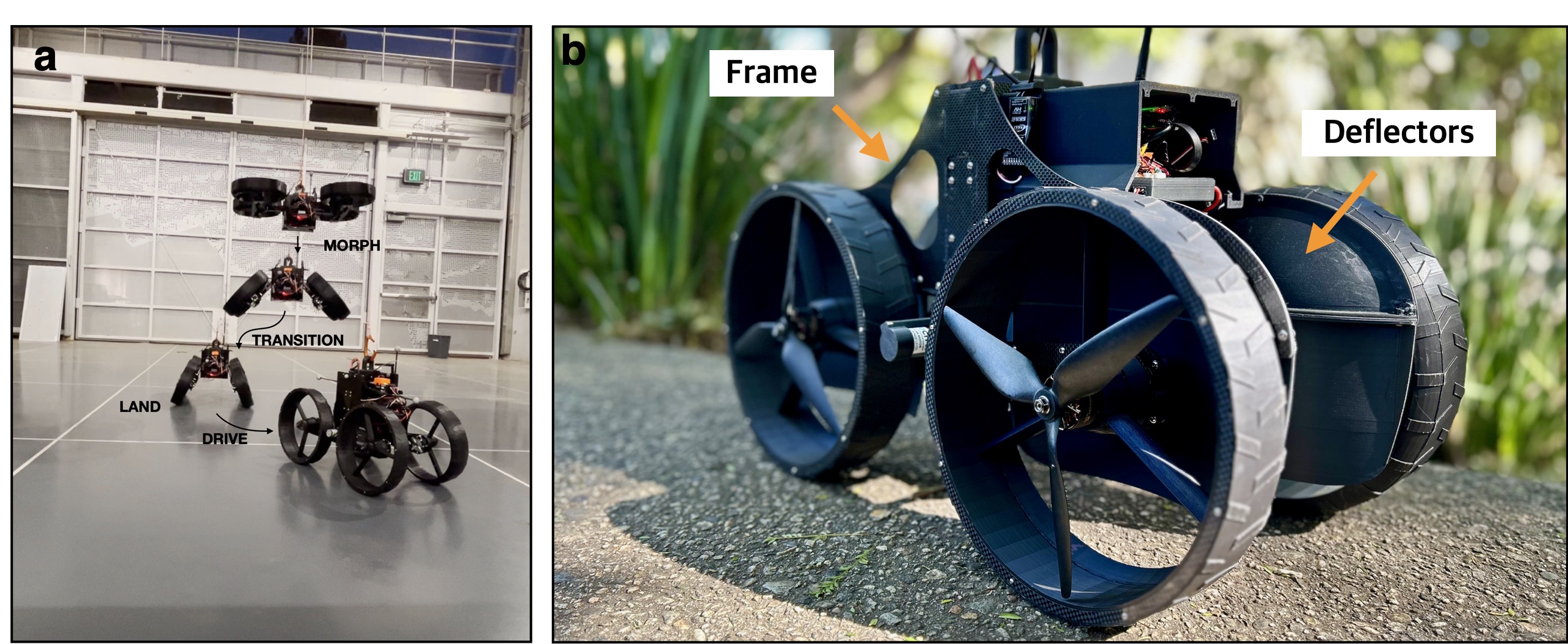}
    \caption{\textbf{Proposed wake vectoring solution.} \textbf{a} Example of a Morphing Aerial Robot, the Aerially Transforming Morphobot (ATMO). This robot is capable of the Morpho-Transition maneuver where the robot transitions from flying to driving in one smooth aerial transformation maneuver. During this maneuver, the rotors are tilted away from the vertical axis resulting in lower thrust available in the vertical direction. reproduced with permission from \cite{Mandralis2025-commeng}. \textbf{b} Our proposed solution is inspired by the flow manipulation techniques used in aircraft design and rocketry. A passive flow deflector is incorporated behind each wheel-thruster combination. This intercepts and redirects the rotor wake, resulting in extended momentum recovery during transition or in transformed configurations.}
    \label{fig:intro}
\end{figure*}

Autonomous robots are poised to revolutionize society, removing the need for humans to perform laborious tasks, improving care of the sick and elderly, and transporting people and goods with greater efficiency and safety. Flying robots promise to serve a unique niche in this vision of the future –– able to reach and manipulate objects and structures where ground robots cannot, while using the airspace to more efficiently navigate and deliver critical materials. 

However, enhanced aerial autonomy requires the ability to operate in cluttered and unpredictable urban environments. In these spaces, current quadrotor technologies are unable to adapt to scenarios such as flying through narrow gaps, perching on walls or ceilings, navigating beneath overhangs, or transitioning between aerial and ground locomotion. Aerial shape change is playing an important role in bridging this gap. Mid-flight reconfiguration has enabled robots to fly through narrow spaces \cite{Mandralis2023,bucki2019,bucki2022,Falanga2019}, perch on surfaces to increase battery life and range \cite{zheng2023metamorphic,Hsiao2023,Hsiao2018}, recover from actuator failures \cite{Mueller2014,Salagame2025-nmpc}, or transition from flying to rolling mode when bi-modal locomotion is needed \cite{Mandralis2025-commeng,Mandralis2025-iros, Sihite2023,Kalantari2020}. 

However, achieving effective aerial shape change remains a significant scientific challenge. Maintaining precise hover during reconfiguration to meet mission requirements demands not only advanced control strategies but often a complete redesign of the control system. This challenge is further compounded by the reduction in available thrust as the propulsion system tilts away from the gravitational axis. For example, the Aerially Transforming Morphobot (ATMO) \cite{Mandralis2025-commeng}, shown in Figure~\ref{fig:intro}(a), re-purposes its thrusters as wheels and employs aerial transformation to transition between flight and drive configurations. Such transformations reduce the vertical thrust available for maintaining position or countering disturbances due to the tilting of the propulsors. Prior work has addressed this limitation through advanced control methods \cite{Mandralis2025-commeng} and deep reinforcement learning \cite{Mandralis2025-iros} to maximize performance, yet physical actuation limits remain the principal constraint.

To solve this problem, we have taken inspiration from the field of aircraft and rocket design where principles of airflow manipulation have long been used to enhance aircraft performance. In rocketry, for example, gimbaled or vectoring nozzles redirect the engine exhaust, which is crucial for steering the vehicle when aerodynamic surfaces are ineffective, e.g. in vacuum \cite{COUNTER1975}. In general, directing engine thrust off-axis provides key benefits including vertical/short takeoff and landing (VTOL/STOL) ability, and significantly higher agility in flight through thrust reversal or vectoring \cite{GOETZ1976,Bare1983}. A notable example is the Harrier Jump Jet, which used a rotating nozzle system to vector its engine exhaust downward – giving it the ability to take off and land vertically on very short runways \cite{Fozard2001-mx}. Other fighter jets likewise use vectored thrust to perform maneuvers that exceed the normal aerodynamic envelope, illustrating how airflow manipulation translates into enhanced maneuverability \cite{BERRIER1978}.

This insight suggests that morphing aerial robots could reap substantial benefits by incorporating flow-re-direction mechanisms into their designs. By reconfiguring their structure or deploying internal deflectors to divert the rotor wake, a morphing drone can recover otherwise-wasted airflow momentum and boost useful thrust. The potential of this approach has been demonstrated in prior work on unmanned aerial vehicles (UAVs). Examples include shape changing platforms that employ movable flaps to enable both multi-rotor and efficient fixed-wing flight \cite{ruben2016,Ruben2016-2,ruben2017,ruben2019}; single-rotor systems equipped with adjustable vanes to redirect thrust without additional actuators \cite{carholt2016}; and quad-rotor vehicles that use deflectors in the propulsive slipstreams to apply lateral forces during hover without altering propeller speeds \cite{henderson2022}. Collectively, these studies highlight the promise of integrating flow manipulation principles into robotic fliers to boost thrust efficiency, enhance control authority, and enable multi-modal flight through mode-switching capability.

In this paper, we translate the aforementioned principles into a novel flow‑deflection mechanism that requires no moving parts or electronics and that is embedded directly in the Aerially Transforming Morphobot's (ATMO) chassis, shown in Figure \ref{fig:intro} (b). As the robot transforms, the deflectors intercept and redirect the rotor wake downward. This mechanism extends momentum recovery in transformed configurations, yielding increased thrust during transitional phases. Our experiments demonstrate a significant vertical thrust augmentation, highlighting the potential of passive fluidic systems in morphing UAV architectures. Remarkably, we found that in extreme aerial configurations where no thrust was available without our deflector design, up to $40\%$ of maximum thrust was recovered, significantly expanding the robot hover capabilities. We believe this innovation opens a promising path toward thrust-efficient, morphing aerial vehicles capable of agile transformations and expanded mission profiles.

\section*{Results}

\begin{figure*}
    \centering
    \includegraphics[width=\linewidth]{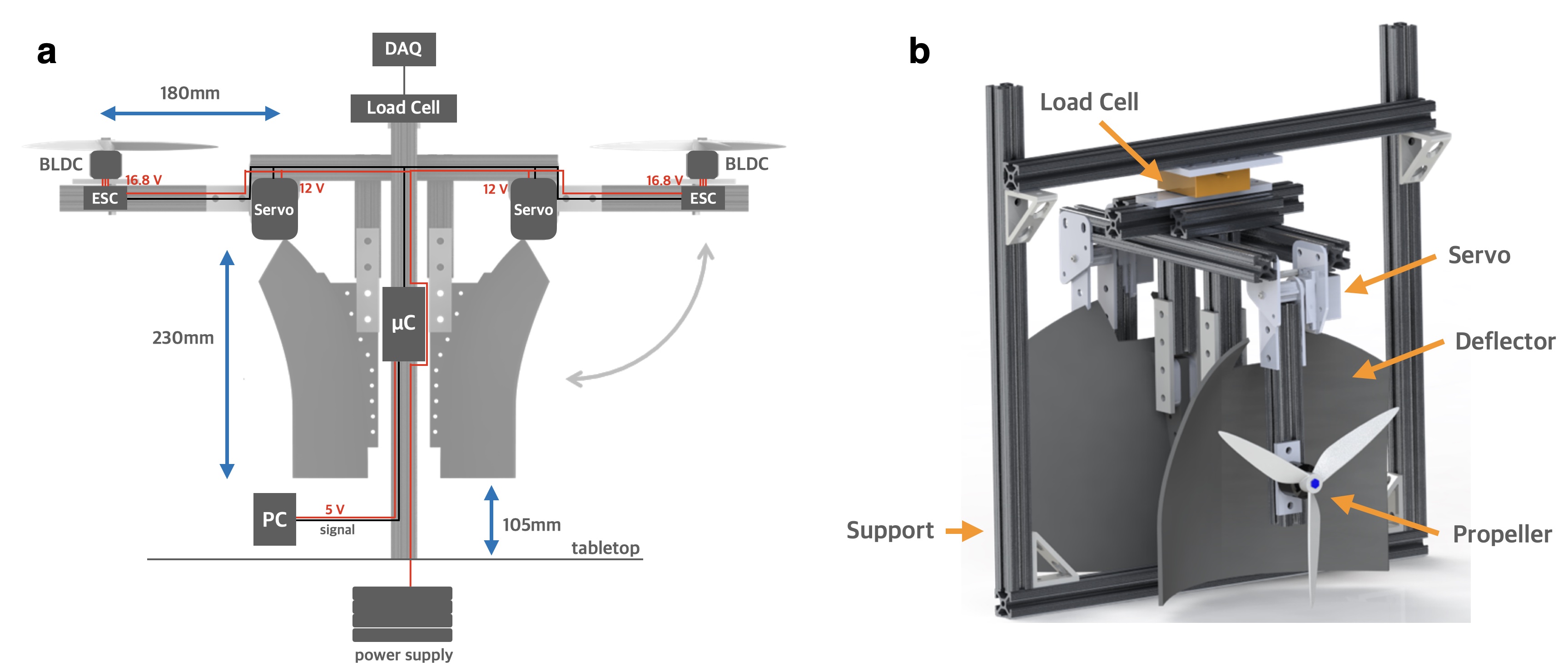}
    \caption{\textbf{Benchtop Experimental Setup.} \textbf{a} Schematic of benchtop test rig with electronic placement, wiring, and dimensions depicted. The thrusters can rotate by servo motors which are controlled from a central microcontroller labeled ($\mu C$). The power supply consists of a 16.8V DC supply that is stepped down to 12V to power the servo motors, and feeds 16.8V to the ESCs which control the rotational speed of two Brushless DC motors (BLDC). A load cell is attached to the top of the test rig and the thrust data is read into a PC using a digital acquisition (DAQ) module. The deflectors are incorporated rigidly onto the main frame. The setup is symmetric to eliminate moment crosstalk interference. \textbf{b} 3D model of the experimental setup. The load cell is depicted, and the propellers are attached to rigid bars that rotate due to a servo motor attached at the pivot point. Everything is mounted onto a rigid frame to eliminate vibrations.
}
    \label{fig:benchtop-setup}
\end{figure*}

\subsection*{Benchtop Experiments}
Benchtop load cell experiments were conducted to evaluate the feasibility of the proposed thrust recovery concept. Inspired by the design of ATMO, a custom test rig was constructed featuring two propellers that direct airflow onto 3D-printed deflector surfaces. The propeller thrust axes can be rotated using two servo motors, replicating the configuration changes that occur during ATMO’s flight-to-drive transition. A load cell is integrated in series with the propeller-deflector assembly to measure the vertical thrust force. To minimize sensor cross-talk and balance the moment acting on the load cell, two identical propeller-deflector assemblies were operated at the same rotational speed. The experimental setup is shown in Figure~\ref{fig:benchtop-setup}(a) and (b).

The deflector was designed as a concave surface with two principal curvatures and sufficient depth to turn the flow effectively without inducing separation. Its area was made larger than that of the propeller disk to ensure full capture of the propeller wake. Particle image velocimetry (PIV) studies have shown that the highest momentum in a propeller flow is concentrated in a ring of fluid near the propeller tips \cite{oshima2023}. Since the force generated on the deflector is directly proportional to the rate at which momentum is redirected vertically, capturing and turning this high-momentum fluid is key to maximizing thrust recovery.

\begin{figure*}[htbp]
    \centering
    \includegraphics[width=1\linewidth]{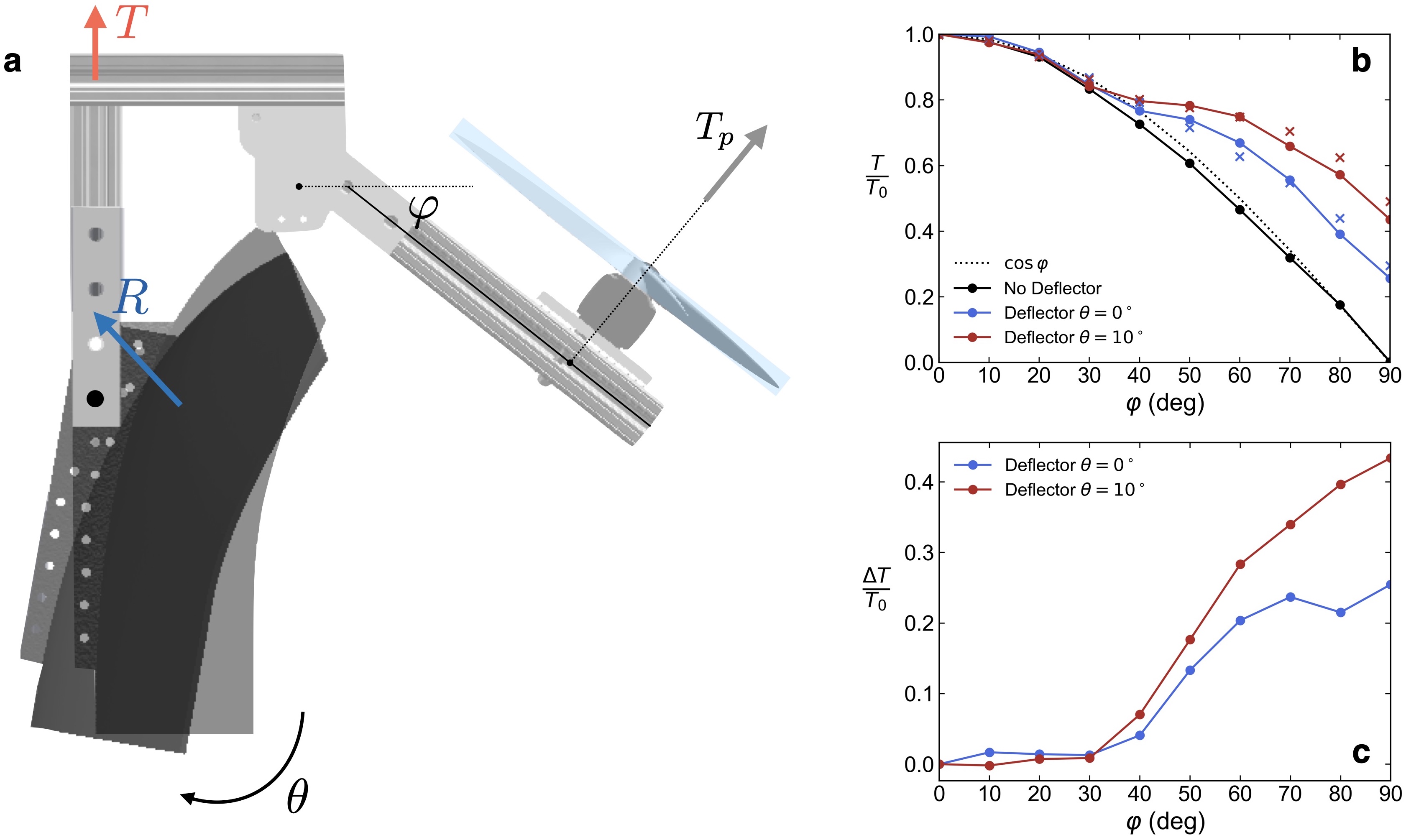}
    \caption{\textbf{Thrust Recovery Results for Benchtop Experiment.} \textbf{a} One half of the benchtop experiment is depicted with key quantities labeled. $\bm R$ is the force acting on the deflector due to the propeller flow, $T_p$ is the thrust produced by the spinning propeller, and $T$ is the overall level of vertical thrust acting on the propeller-deflector assembly. $\theta$ is the rotation of the deflector from the baseline configuration where the deflector is pointing vertically down, and $\varphi$ is the tilt angle of the propeller thrust axis. \textbf{b} Thrust values from the experimental setup as a function of $\varphi$ for two different deflector angles $\theta\in\{0^\circ,10^\circ\}$ as a ratio of the thrust at $\varphi=0^\circ$. The thrust with no deflectors in place is plotted in black showing close agreement with the theoretical $\cos\varphi$ decay curve. The result of the numerical flow simulations are overlayed with x markers over the $\theta=10^\circ$ and $\theta=20^\circ$ cases, showing reasonable agreement. \textbf{c} Thrust recovery as a ratio of the thrust level at $\varphi=0^\circ$ for two different deflector exit angles.}
    \label{fig:benchtop-results}
\end{figure*}

We first oriented the deflector exit angle parallel to gravity ($\theta = 0^\circ$) and measured the total force at the load cell for ten different propeller tilt angles ($\varphi$), ranging from $\varphi = 0^\circ$ (thrust axis vertical) to $\varphi = 90^\circ$ (thrust axis horizontal). The experimental setup is shown in Figure~\ref{fig:benchtop-results}(a). The resulting thrust, normalized by the thrust at $\varphi = 0^\circ$ ($T_0$), is plotted as $T/T_0$ versus $\varphi$ in Figure~\ref{fig:benchtop-results}(b).

As expected, the thrust generally decreased with increasing tilt angle as the propeller was oriented away from vertical. When compared to tests without deflectors (also shown in Figure~\ref{fig:benchtop-results}(b)), the deflector provided markedly better thrust recovery at tilt angles greater than $40^\circ$. For reference, we compared our measurements to the theoretical decay $T/T_0 = \cos\varphi$, which represents the loss of vertical thrust component due purely to geometric projection, absent aerodynamic effects.

With $\theta = 0^\circ$, the deflector had negligible influence on thrust for $\varphi < 40^\circ$. However, beyond this angle, the thrust no longer followed the cosine trend, and a substantial recovery was observed. At $\varphi = 90^\circ$, where thrust would otherwise fall to zero, approximately $25\%$ of $T_0$ was retained due to the deflector. This excess thrust is highlighted in Figure~\ref{fig:benchtop-results}(c), which plots $\Delta T / T_0$, the recovered thrust beyond the cosine prediction.

To explore the potential for enhanced recovery, we repeated the experiment with the deflector exit angle set to $\theta = 10^\circ$. The angled deflector yielded significantly better performance at high tilt angles. Thrust levels were similar to the $\theta = 0^\circ$ case below $\varphi = 40^\circ$, but consistently higher at larger angles. In particular, at $\varphi = 90^\circ$, the thrust recovery exceeded $40\%$ of $T_0$. Due to mechanical constraints of the test bench, we were unable to test exit angles greater than $\theta = 10^\circ$.

\subsection*{Effect of Deflector Angle on Thrust Recovery}

\begin{figure*}[thbp]
    \centering
    \includegraphics[width=1\linewidth]{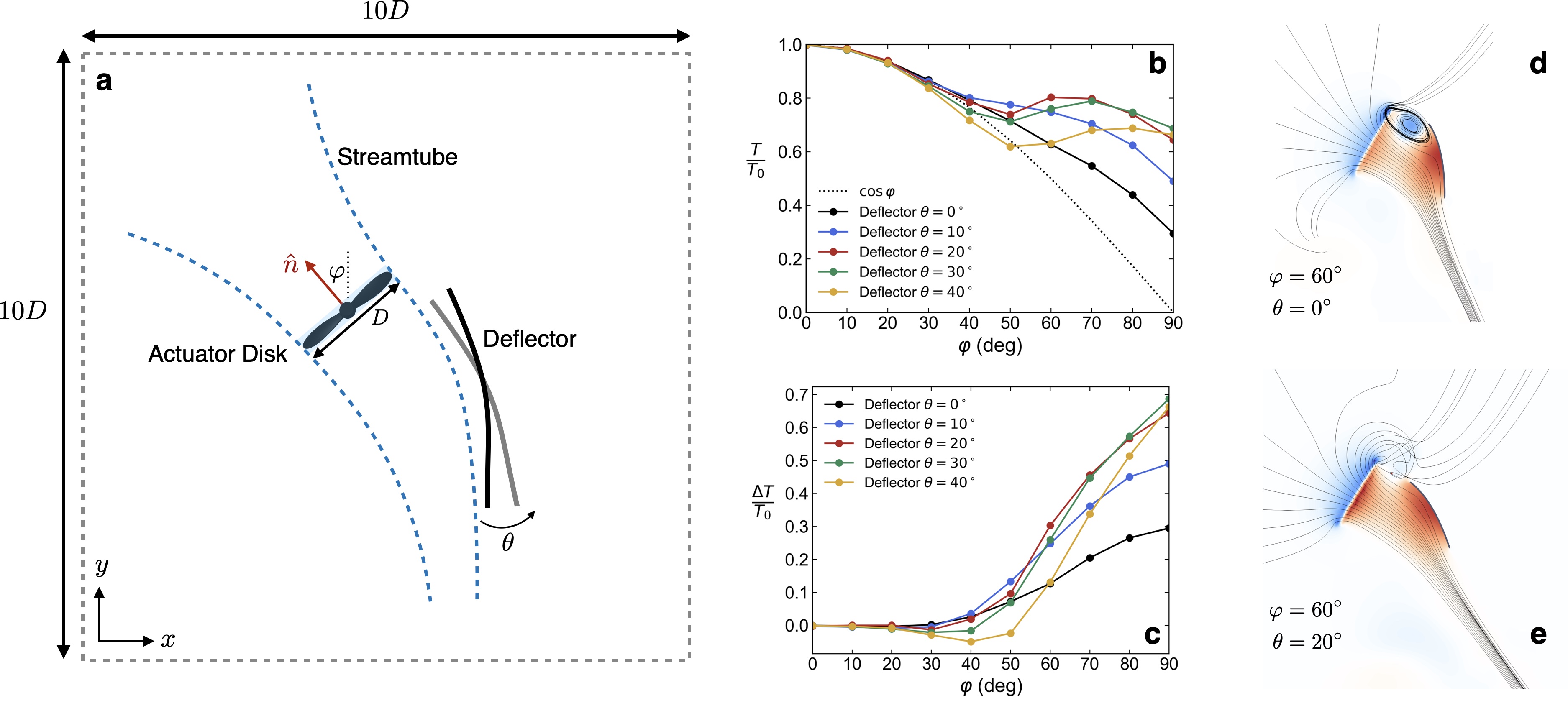}
    \caption{\textbf{Influence of Deflector Exit Angle on Thrust Recovery.} \textbf{a} Schematic of the computational domain used for numerical flow simulations. A two-dimensional slice of the three-dimensional domain is depicted. The actuator disk is a cell zone of diameter equal to the propeller diameter $D$ that is tilted at an angle $\varphi$ from the vertical axis. A 2D slice of the deflector and the deflector exit angle $\theta$ is depicted. \textbf{b} Thrust values as a ratio of the thrust at $\varphi=0^\circ$ are plotted for 5 different exit angles  $\theta\in \{0^\circ,10^\circ,20^\circ,30^\circ,40^\circ\}$. \textbf{c} Thrust recovery  values for the same exit angles. \textbf{d,e} Representative pressure fields for two cases are shown. The streamlines from the 3D velocity field are projected onto the two-dimensional slice. }
    \label{fig:numerical-results}
\end{figure*}

To further investigate the effect of deflector exit angle on thrust recovery performance, we conducted numerical simulations replicating the benchtop experimental setup. Using a simplified model of the propeller as an actuator disk \cite{rankine1865mechanical,froude1889part} we performed Reynolds-Averaged Navier-Stokes (RANS) simulations using an open-source finite-volume solver, OpenFOAM \cite{jasak2007openfoam,jasak2009openfoam}. This allowed us to simulate propeller-deflector cases with exit angles past $\theta=10^\circ$. The computational domain is shown schematically in Figure \ref{fig:numerical-results} (a). Further details of the numerical solver and setup are provided in the Methods section.

We first validated the numerical simulations against the experimental data from the previous section. Good agreement with the experimental measurements was observed for the $\theta\in\{0^\circ,10^\circ\}$ cases, as can be seen in Figure \ref{fig:benchtop-results} (b) where the numerical predictions are overlayed on the experimental values. Small discrepancies in thrust levels may be attributed to factors such as discretization errors or the omission of mechanical components of the test rig in the model, which could influence the flow. 

We then examined the thrust recovery performance for deflector angles greater than $\theta = 10^\circ$. The results, presented in Figure \ref{fig:numerical-results} (b) and (c), indicate that thrust recovery generally improved as the deflector angle increased, up to $\theta=30^\circ$ for high propeller tilt angles ($60^\circ\leq\varphi\leq90^\circ$). The optimal deflector angle for these high tilt angles lies between $\theta=20^\circ$ and $\theta=30^\circ$. At $\theta=20^\circ$, for example, peak thrust recovery of $0.8$ was observed at $60^\circ$ – a substantial improvement over the baseline case with no deflector tilt. Further increases beyond $\theta = 30^\circ$ did not yield additional benefits; instead, thrust recovery decreased across all tilt angles, suggesting that deflector tilting enhances performance only up to a critical angle, beyond which it becomes detrimental.

To understand the physical mechanisms driving this behavior, we analyzed the pressure fields and representative streamlines of the various deflector configurations. When the deflector was vertical ($\theta = 0^\circ$), part of the flow recirculated near the deflector entrance, creating a low-pressure region that caused a loss of momentum as flow escaped around the deflector. This is illustrated in Figure~\ref{fig:numerical-results}(d), which shows the flow field and streamlines for $\varphi = 60^\circ$ and $\theta = 0^\circ$. The low-pressure and recirculation regions are clearly visible in a two-dimensional slice through the center of the deflector and actuator disk. In contrast, at $\theta = 20^\circ$, these recirculation zones disappeared (Figure~\ref{fig:numerical-results}(e)), resulting in uniformly high pressure beneath the deflector and more momentum being redirected downward. This produced greater overall thrust and underscores the importance of aerodynamic optimization for maximizing thrust recovery across all propeller tilt angles.

\subsection*{Implementation of Passive Flow Deflectors on ATMO}

\begin{figure*}[thbp]
    \centering
    \includegraphics[width=1\linewidth]{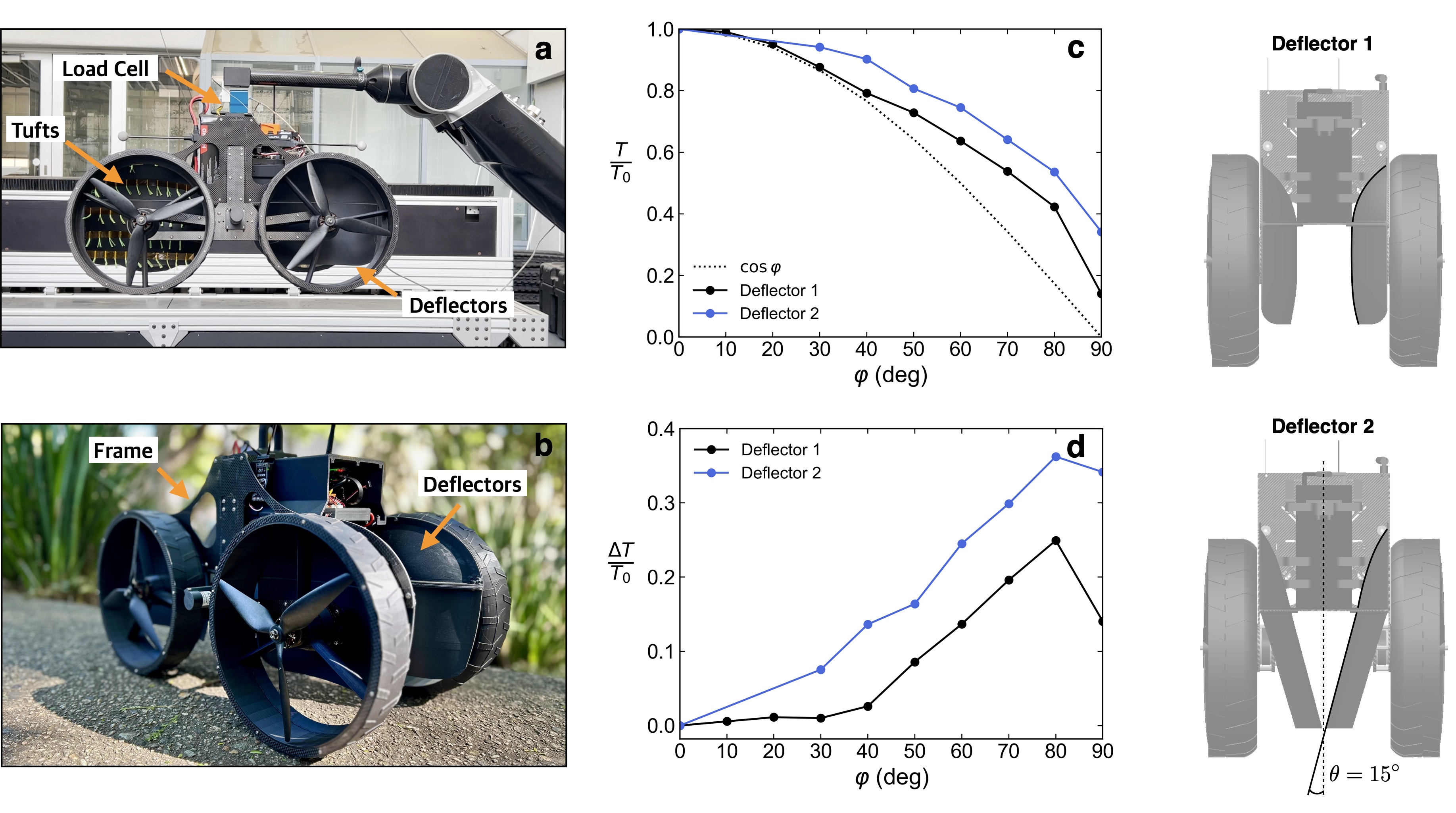}
    \caption{\textbf{Implementation of Wake Vectoring Method on Aerially Transforming Morphobot (ATMO).} \textbf{a} Load cell test rig. ATMO is mounted on a robotic arm with a load-cell that measures the vertical force. Tufts have been included on one of the deflectors for some visual feedback of the flow field near the deflector surface. \textbf{b} The deflectors implemented on ATMO. \textbf{c} Thrust values as a ratio of the thrust at $\varphi=0^\circ$ for both deflector designs compared to the $\cos\varphi$ decay curve. \textbf{d} Thrust recovery values for both deflectors. (right) The two deflector design are shown from the frontal perspective. The Deflector 2 design implements insight from benchtop experiments and numerical flow simulations by maximizing the exit angle of the deflector, achieving superior performance.}
    \label{fig:atmo-experiments}
\end{figure*}

We applied the principles from the benchtop experiments and numerical simulations to design deflectors that could be integrated directly into ATMO’s chassis. Two design cases were considered. The first was a baseline deflector with no deflector exit angle relative to the vertical axis. This deflector had a width equal to the propeller disk area, aiming to redirect the maximum flow with the smallest possible footprint. Its length was kept smaller than the wheel diameter to maintain sufficient ground clearance, ensuring ATMO could still operate in drive mode when the wheels are fully retracted. The deflector was designed to be as deep as possible without interfering with the onboard robot electronics. 

The second design incorporated insights from our earlier experiments and simulations by adopting the largest feasible exit angle. Due to ATMO’s short wheelbase in drive mode, the maximum exit angle achievable was $15^\circ$. The deflector was also widened and lengthened beyond the propeller disk area to capture and redirect more of the high-momentum flow near the propeller tips. The two deflector designs are shown in Figure~\ref{fig:atmo-experiments}; Figure~\ref{fig:atmo-experiments}(b) provides a photograph of Deflector 1 integrated onto ATMO in its drive configuration.

Both deflector designs were mounted on ATMO, which was secured to a single-axis load cell as shown in Figure~\ref{fig:atmo-experiments}(a). We measured the total vertical force at ten propeller tilt angles between $\varphi = 0^\circ$ and $\varphi = 90^\circ$. The results, presented in Figure~\ref{fig:atmo-experiments}(c) and (d), show that both designs improved thrust recovery at high tilt angles. As anticipated, Deflector 2 outperformed Deflector 1, achieving a peak thrust recovery just under $40\%$ at $\varphi = 80^\circ$, with only a small decline at $\varphi = 90^\circ$. The use of a $15^\circ$ exit angle consistently enhanced thrust across all tilt angles, supporting our earlier hypothesis that increasing deflector angle improves thrust recovery efficiency.

Interestingly, Deflector 2 also showed greater thrust recovery at low tilt angles ($\varphi < 40^\circ$) compared to the benchtop results. This may be due to aerodynamic interactions between the propeller flow and the surrounding wheels, or the specific geometry of the deflector on ATMO. These results confirm that passive flow deflectors can be effectively integrated into morphing aerial-ground robots, providing significant aerodynamic benefits without active control mechanisms.

\subsection*{Hover in Extreme Aerial Configurations}

\begin{figure*}[thbp]
    \centering
    \includegraphics[width=1\linewidth]{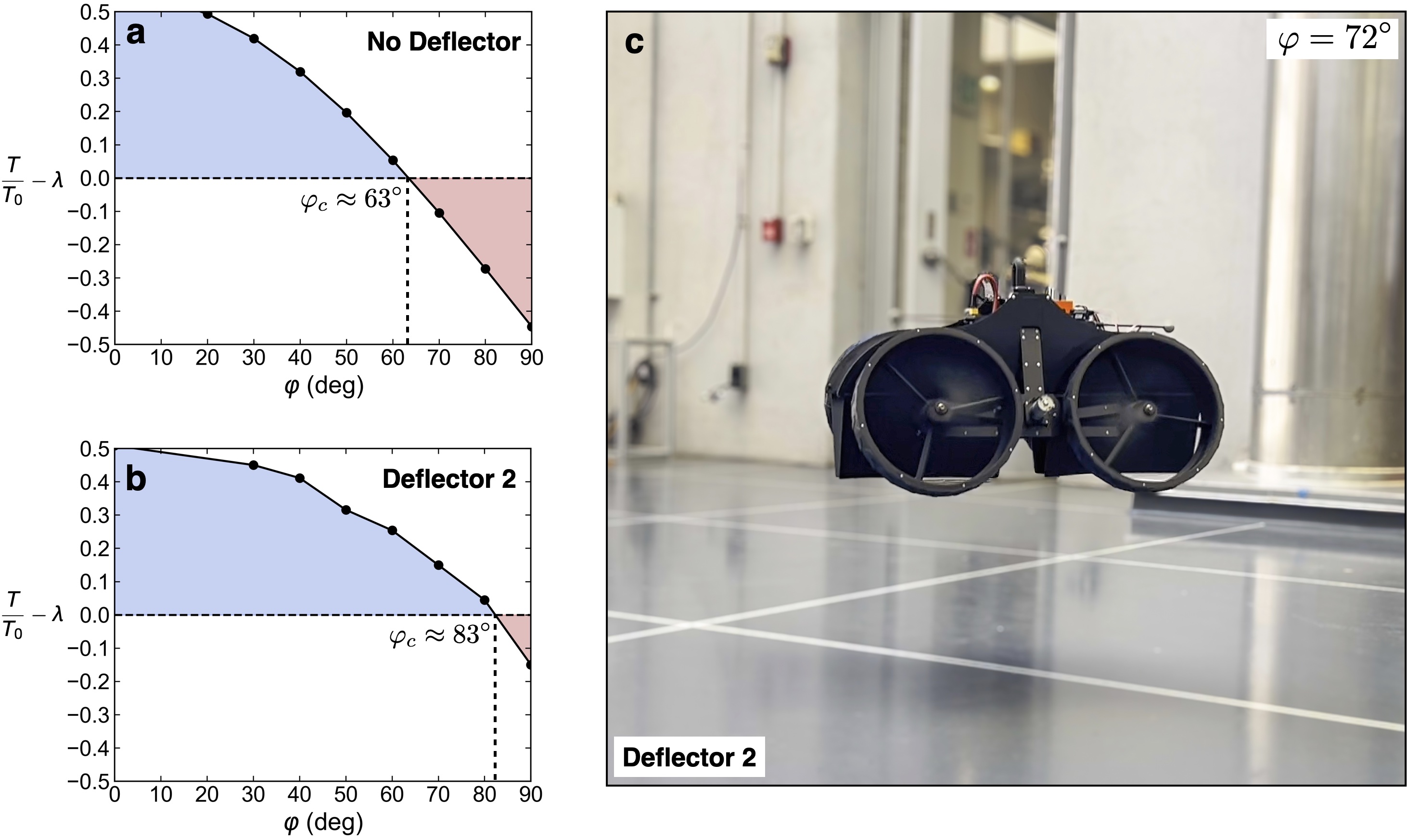}
    \caption{\textbf{ATMO Hovering in an Extreme Aerial Configuration.} \textbf{a} The excess hover thrust, $\frac{T}{T_0}-\lambda$, is plotted as a function of the body tilt angle $\varphi$ for the case of no deflectors; $\lambda$ is the reciprocal thrust to weight ratio. Without deflectors the robot weighs 5 kg so $\lambda=0.45$, where the experimentally measured maximum thrust of 11.2 kg was used. \textbf{b} Excess hover thrust with Deflector 2 incorporated on ATMO. Since the deflector adds 0.5 kg of weight, $\lambda=0.49$. Despite the increased weight, the critical hover angle still increases significantly; $\varphi_c = 83^\circ$. \textbf{c} ATMO hovering with $\varphi=72^\circ$ in an extreme configuration in Caltech's Center for Autonomous Systems and Technology (CAST) flight arena.
    }
    \label{fig:hover}
\end{figure*}

The thrust required for hover depends on both the vehicle’s weight and its body configuration. At each tilt angle $\varphi$ the maximum thrust available is given by the thrust recovery curve as $T_{\max}(\varphi) = \frac{T(\varphi)}{T_0}T_{\max}$, where $T(\varphi)/T_0$ is the normalized thrust recovery curve. For a vehicle of mass $m$, the maximum tilt angle at which hover is possible corresponds to the intersection of this curve with the required hover thrust:
\begin{equation}
\frac{T(\varphi)}{T_0} = \frac{m g}{T_{\max}} \equiv \lambda, \label{eq:hover}
\end{equation}
where $\lambda$ is the reciprocal of the vehicle’s thrust-to-weight ratio. For ATMO without deflectors $\lambda = 0.45$; with thrust deflectors $\lambda = 0.49$.

The solution to Eq. \ref{eq:hover} is illustrated in Figure \ref{fig:atmo-experiments}(a) and (b) for the cases without deflectors and with Deflector 2, respectively. Hover is achievable at tilt angles to the left of the intersection between the horizontal line at $\lambda$ and the thrust recovery curve; the available control authority (excess thrust) is highlighted in blue. Beyond this intersection, thrust is insufficient for hover and control, as indicated in red. For Deflector 2, the critical hover angle is $\varphi_c = 83^\circ$, representing a substantial $20^\circ$ increase compared to the configuration without deflectors.

We experimentally tested ATMO’s ability to hover in extreme configurations equipped with Deflector 2. Hover was successfully achieved at $\varphi = 72^\circ$ while maintaining stability and compensating for disturbances. A snapshot of the experiment is shown in Figure~\ref{fig:atmo-experiments}(c) and in Supplementary Video 1. The robot was able to take off from the drive configuration, sustain hover, and perform controlled hops to new locations. Beyond $\varphi = 72^\circ$, ATMO could initiate takeoff but lacked sufficient control authority to stabilize vertical position against disturbances. This phenomenon, consistent with prior observations for ATMO without deflectors \cite{Mandralis2025-commeng}, reflects a general limitation of systems operating near actuator saturation.

Remarkably, the same control architecture used for conventional quadrotor flight—with only minor gain tuning—was able to stabilize ATMO even in these extreme aerial configurations. This contrasts sharply with the deflector-free case, where specialized control strategies were required to manage the horizontal thrust components induced by rotor tilting \cite{Mandralis2025-commeng}. The deflectors not only augmented vertical thrust but also simplified control by redirecting momentum downward and reducing horizontal force components. Thus, passive flow deflectors provide a dual benefit: enhanced thrust recovery and reduced control complexity.

\section*{Discussion}
We have presented an aerodynamics-inspired method for enhancing the thrust efficiency of aerial robots that reconfigure mid-flight. Our findings demonstrate that passive wake vectoring can provide substantial thrust recovery during morphing flight, with up to 40\% of vertical thrust regained in configurations where no vertical thrust would be expected without flow manipulation. These results highlight that carefully designed flow-deflection surfaces can mitigate thrust losses associated with mid-flight reconfiguration, reducing reliance on additional actuators or complex control strategies. Compared to prior solutions which used active thrust vectoring through model-predictive control schemes or reinforcement learning \cite{Mandralis2025-commeng,Mandralis2025-iros}, passive wake vectoring offers a low-complexity alternative that does not require additional electronics or actuation.

The ability to passively steer rotor wake opens new design opportunities for morphing aerial robots. We successfully implemented the deflector design on ATMO to hover at extreme body angles, indicating that shape-changing aerial robots equipped with passive wake vectoring can achieve more aggressive transformations without sacrificing control authority. This expands the operational envelope, particularly for applications in cluttered environments, where drones must rapidly adapt their morphology to navigate tight spaces or interact with structures. For example, operating ATMO at $\varphi=72^\circ$ tilt angle reduces its cross-sectional width by 20~cm, enabling entry into narrow gaps where flight or driving can continue.

This study also provides insight into the physical principles governing thrust recovery through passive flow deflection. Key design aspects identified as critical for effective thrust recovery include: (i) ensuring the deflecting surface fully captures the propeller disk area, and (ii) preventing flow separation or recirculation on the deflector surface to maximize momentum redirection. We further developed a numerical simulation framework that can be readily adapted to other case studies. By modeling the propeller as an actuator disk and applying a simple momentum source term to the velocity transport equations, the framework avoids the need for complex rotating meshes while yielding simulation results in reasonable agreement with experimental measurements. The simulation framework is available online (see Code Availability section).

While promising, this approach has limitations. The level of thrust recovery achieved is highly dependent on the geometry of the deflector and the specific morphing configuration. Although the design principles outlined here can inform future designs, the current deflector geometry will not necessarily generalize to other morphing aerial robots without modification.

Future work will explore how passive wake deflection can be integrated at the robot design stage to optimize thrust recovery across a broader range of configurations. Leveraging pre-existing structural elements for wake deflection could minimize the added weight and form drag associated with dedicated deflectors. Finally, augmenting passive wake vectoring with active flow control technologies, such as movable vanes, may further expand the capabilities of morphing aerial robots.

\section*{Methods}

\subsection*{Numerical Simulations}
For the fluid simulations we used the open-source computational fluid dynamics software OpenFOAM \cite{jasak2007openfoam,jasak2009openfoam} and the algorithm \verb|simpleFOAM| to solve the incompressible, Reynolds Averaged Navier-Stokes equations,
\begin{align*}
\frac{\partial\bar u_i}{\partial x_i} &= 0\\
    \rho \, \bar{u}_j \frac{\partial \bar{u}_i}{\partial x_j}
&= \rho \, \bar{a}_i
+ \frac{\partial}{\partial x_j}
\left[
- \bar{p} \, \delta_{ij}
+ \mu \left(
\frac{\partial \bar{u}_i}{\partial x_j}
+ \frac{\partial \bar{u}_j}{\partial x_i}
\right)
- \rho \, \overline{u_i' u_j'}
\right],
\end{align*}
for $\bm {\bar u}$ and $\bar p$ which are the time-averaged velocity and pressure fields respectively. $\rho$ is the density and $\mu$ is the viscosity of air at $15^\circ$~C. The Einstein index notation of summation over repeated indices is implied. $\bm a$ is a body acceleration source that we impose to simulate the propeller flow. The RANS equations are closed by modeling the Reynolds Stresses using Menter's Shear Stress Transport (SST) $k$-$\omega$ model \cite{Menter1993,Menter1994}. 

We used a cubic simulation domain with edge length $2$ meters which corresponds to $10$ times the propeller diameter ensuring that the boundaries do not affect the flow. The propeller and deflector were placed in the center of the simulation domain with relative distances determined by the benchtop geometry shown in Figure \ref{fig:benchtop-setup}(a) as well as the angles $\varphi,\theta$ for each case. The mesh consisted of $2.7$ million hexahedral cells and was adjusted to the deflector surface using OpenFOAM's built in mesh snapping algorithm \verb|snappyHexMesh|. The mesh was refined around the deflector surface to ensure that the surface was captured accurately and did not contain any unwanted surface imperfections. 

To simulate the propeller without needing a rotating mesh or the computational costs of  a higher-fidelity simulation, we resorted to a simplified model based on actuator disk theory –– one of the simplest and oldest mathematical tools for modeling screw propellers \cite{VANKUIK20151}. The key simplification made by this theory is to replace the load on the real propeller by a uniform and normal pressure distribution on an infinitely thin, permeable disc \cite{rankine1865mechanical,froude1889part}. To implement this in OpenFOAM, we defined a cylindrical zone centered at the propeller of height $10$mm and radius equal to the propeller radius. In this cell zone we imposed a source term in the momentum equation of $T_p=12 N$ per cubic meter, simulating the effect of a propeller pushing air through the propeller disk. 

The no-slip boundary condition was applied on the deflector surface and a \verb|pressureInletOutletVelocity| boundary condition was applied to the six edges of the cubic domain which adjusts the inflow or outflow depending on the local pressure field allowing the boundary to switch between inlet and outlet behavior based on the solution during the simulation. 

The force on the deflector was computed by integrating the stress tensor $\bm \sigma$ at the deflector surface,
\begin{align*}
    \bm R = \int_S \bm \sigma \cdot \bm {n} \ \textrm{d} S,
\end{align*}
where $\bm \sigma = - p \bm I + \bm \tau$ and $\bm \tau$ is the viscous stress tensor. To simulate the cases where the deflector was tilted, we kept the deflector fixed in the mesh local coordinates and transformed the position of the actuator disk and thrust force applied relative to the deflector using the two dimensional rotation matrix 
\begin{equation*}
\bm Q_z(\theta) = \begin{bmatrix}
    \cos\theta & -\sin\theta \\ \sin\theta & \cos\theta
\end{bmatrix}.
\end{equation*}
For the cases where $\theta$ was non zero, the deflector force $\bm R$ was first computed in the transformed coordinates and was transformed back to lab coordinates as follows, 
\begin{equation*}
    T = T_p\cos(\varphi) + \bm Q_z(\theta)\bm R \cdot \bm e_y,
\end{equation*}
where $T_p$ is the thrust (momentum source) imposed on the actuator disk and $T$ is the total vertical foce on the propeller-deflector assembly. 

We ran all simulation cases in parallel using 12 CPU cores. Each simulation was terminated when all residuals had dropped below a predefined tolerance and took an average of 15 minutes to run to convergence on a Desktop Computer equipped with an AMD Ryzen 9 7900X processor. The code used is publicly available (See Code Availability section). 

\subsection*{Benchtop Experiment}
The experimental setup is shown in Figure~\ref{fig:benchtop-setup}(a) and (b). Two counter-rotating propellers were used to minimize the net moment acting on the load cell. The entire assembly was mounted on an adjustable aluminum frame, with the load cell positioned between the frame and the motor–propeller assemblies. Each motor was powered by a dedicated 4S LiPo battery, providing 16.8~V when fully charged. To ensure consistent performance, the batteries were recharged after each test set to compensate for voltage drop over successive runs.

Tests were performed at various rotational speeds and tilt angles, with the angle incremented in 10° steps from 0° (horizontal) to 90° (vertical). The tilt angle was controlled via an Arduino microcontroller, which actuated the servo motors. Motor speed was regulated using APD80F3 ESCs (Advanced Power Drives), with the Cine66 KV1125 brushless motors (T-Motor) driving 9-inch (228~mm) HQ-90503 three-blade propellers.

A 3A60A three-axis load cell (Interface), capable of measuring forces up to 50N per axis, was used to record thrust. Data were acquired at 250Hz using the BlueDAQ software (Interface). Each measurement consisted of setting the propellers to the target RPM and recording the thrust for 15 seconds. The data were analyzed to compute the mean and standard deviation of thrust for each run. To avoid ground effect influences on the measurements, all tests were conducted with the deflectors positioned at least 2m from the nearest surface.

\subsection*{System Integration Load Cell Tests}
ATMO, equipped with the deflectors, was mounted onto a robotic arm that positioned the vehicle sufficiently far from the ground to eliminate ground effect. A single-axis load cell was integrated in series between ATMO and the robotic arm to measure vertical thrust. The orientation of ATMO relative to the ground was set in advance using a digital angle gauge and level meter. Power was supplied by a 24V source capable of delivering up to 100A of continuous current. For each deflector design and tilt angle tested, the propellers were operated at 40\% of their maximum speed, and thrust was recorded for 15~seconds using the same BlueDAQ software and data acquisition hardware as in the benchtop experiments.

\subsection*{Robot Hardware Description}
The Aerially Transforming Morphobot (ATMO) is described in detail in \cite{Mandralis2025-commeng}. We provide here a brief summary of key hardware aspects relevant to this work.



The deflectors are 3D-printed using a Bambu 3D printer in PLA, each weighing about 0.1 kg. To integrate them into the chassis, custom carbon fiber plates were designed and CNC-cut, and the deflectors were mounted onto the chassis using screws. The total additional weight due to the deflectors is 0.5 kg. The total thrust produced by each propeller is 2.8 kg. The weight of the robot without deflectors was 5 kg and with deflectors it was 5.5 kg. This results in a thrust to weight ratio of 2.23 without deflectors and 2.04 with deflectors; equivalently $\lambda=0.45$ without deflectors and $\lambda=0.49$ with deflectors.

\subsection*{Hover experiments}
ATMO, equipped with Deflector 2, was configured at a tilt angle of $\varphi = 72^\circ$. The vehicle was powered by a 2400mAh 6S LiPo battery, and the total weight, including the deflectors, was 5~kg. A CubeOrange flight controller running PX4 Autopilot software \cite{Meier2015} provided onboard control. The body rate controller gains were only slightly adjusted to enhance performance, while the attitude control gains remained at the default manufacturer settings. ATMO was operated in stabilize mode, in which the control inputs correspond to desired roll, pitch, and yaw Euler angles, and the onboard controller stabilizes these through nested attitude and body rate control loops. The control allocation (actuator mixing) was left unmodified.

\section*{Data availability}
Data will be provided upon request.

\section*{Code availability}
Code for the numerical simulations is publicly available \href{https://github.com/mandralis/DeflectorWithActuatorDisk}{here}.

\bibliographystyle{ieeetr}
\bibliography{main.bib}

\section*{Acknowledgements}
We thank Richard M. Murray for numerous insightful discussions and advice on the conception and execution of this project. Joshua Gurovich and Jack Caldwell provided help with 3D printing and cutting the Carbon Fiber plates. Brandon Nilles provided his assistance when adapting the initial design of the deflectors to ATMO's chassis. We also thank Sean Devey and Scott Bollt for discussions on the fluid dynamics of thrust recovery. This work was supported by funding from the Center for Autonomous Systems and Technology. Ioannis Mandralis is an Onassis Scholar and was supported by the GALCIT graduate student endowment fellowship.  

\section*{Author contributions}
I.M. and M.G. conceived the research idea and designed the experiments; S.S and I.M. designed the initial deflector and performed the benchtop test rig experiments; I.M. performed the system integration experiments, numerical simulations, analyzed the data, and wrote the initial manuscript draft with input from M.G.; All authors discussed the results and reviewed the final manuscript.

\section*{Competing interests}
The authors declare they have no competing interests. 

\end{document}